\documentclass{article}
\usepackage{spconf,amsmath,graphicx}
\usepackage{url}
\usepackage{multirow}
\usepackage{amssymb}

\title{APB2FaceV2: Real-Time Audio-Guided Multi-Face Reenactment}
\name{Jiangning Zhang\(^1\), Xianfang Zeng\(^1\), Chao Xu\(^1\), Jun Chen\(^1\), Yong Liu\(^1\), Yunliang Jiang\(^2\)}
\address{\(^1\)Zhejiang University, Hangzhou, Zhejiang, China \\
\(^2\)Huzhou University, Huzhou, Zhejiang, China}

\begin{document}
%
\maketitle
%
\begin{abstract}
  Audio-guided face reenactment aims to generate a photorealistic face that has matched facial expression with the input audio. However, current methods can only reenact a special person once the model is trained or need extra operations such as 3D rendering and image post-fusion on the premise of generating vivid faces. To solve the above challenge, we propose a novel \emph{R}eal-time \emph{A}udio-guided \emph{M}ulti-face reenactment approach named \emph{APB2FaceV2}, which can reenact different target faces among multiple persons with corresponding reference face and drive audio signal as inputs. Enabling the model to be trained end-to-end and have a faster speed, we design a novel module named Adaptive Convolution (AdaConv) to infuse audio information into the network, as well as adopt a lightweight network as our backbone so that the network can run in real time on CPU and GPU. Comparison experiments prove the superiority of our approach than existing state-of-the-art methods, and further experiments demonstrate that our method is efficient and flexible for practical applications\footnote{\url{https://github.com/zhangzjn/APB2FaceV2}}.
\end{abstract}

\section{Introduction}
\label{sec:intro}
  Audio-guided face reenactment is a task to generate photorealistic face images under the condition of audio input, which has promising applications such as animation production, virtual announcer, and game. In this paper, different from current methods that can only reenact a special person once the model is trained, we focus on a more challenging task: audio-guided multi-face reenactment, where different target faces among several persons can be reenacted using one unified model.

  Benefited from the development of neural network, many methods have achieved good results in the audio-to-face task. Cudeiro~\textit{et al.}~\cite{cudeiro2019capture} present a speech driven facial animation framework named VOCA that can fully automatically outputs a realistic character animation given a speech signal and a static character mesh. Works~\cite{yi2020audio, Audio2face} employ the LSTM~\cite{LSTM} model to generate orofacial movement from acoustic features for predefined 3D model. Though 3D model-based methods can obtain vivid results, they need high costs for hardware and predefined 3D model as well as post-processing time consumption. Thus the pixel-based method is born to conduct the audio-to-face task. Duarte~\textit{et al.}~\cite{Wav2Pix} propose the Wav2Pix to generate the face image by an encoded audio vector in an adversarial manner, while Zhang~\textit{et al.}~\cite{APB2Face} design an APB2Face model that immensely improves the quality of the generated image. Consistent with pixel-based method, we design our model in a generative adversarial manner that can reenact photorealistic images and is easy to follow.

  However, almost all of the current approaches~\cite{chung2017you, Wav2Pix, APB2Face} can only reenact one special person once the model is trained on the premise of generating vivid faces, meaning they are not competent to reenact various faces using a unified model. In order to solve the above problem, we propose a novel APB2FaceV2 to reenact different target faces among multiple persons with corresponding reference face and drive audio information, which has more practical application value. Specifically, the proposed APB2FaceV2 consists of an Audio-aware Fuser that extracts embedded geometric vector from input audio, head pose, and eye blink information, as well as a Multi-face Reenactor that generates target faces with a reference face and the geometric vector. At the same time, we find that nearly all current approaches do not take the model size into account that is important for practical applications. So we come up an Adaptive Convolution (AdaConv) to infuse audio information into the network so that the model can be trained in an end-to-end manner, as well as employ a modified lightweight network~\cite{Autogan-distiller} as our backbone so that the model can run in real time.
  Specifically, we make the following four contributions:

  \romannumeral1) A novel \emph{APB2FaceV2} is proposed to to reenact different target faces among multiple persons using one unified model.

  \romannumeral2) We design a new vector-based information injection module named \emph{AdaConv} that achieves an end-to-end training.

  \romannumeral3) A lightweight backbone is adopted so that the method can run on CPU or GPU in real time.

  \romannumeral4) Experimental results demonstrate the efficiency and flexibility of our proposed approach.

\begin{figure*}[!ht]
  \centering
  \includegraphics[width=1.9\columnwidth]{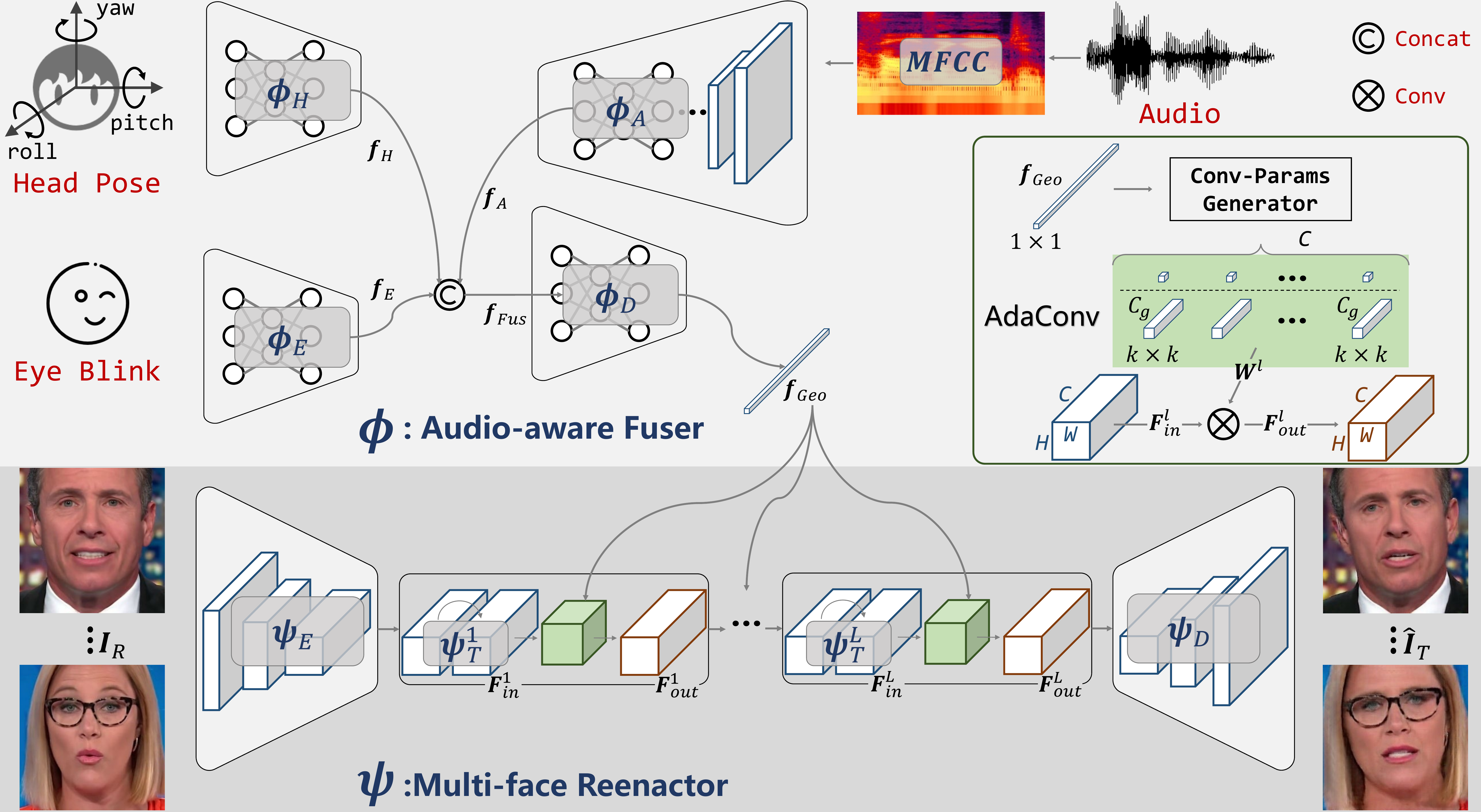}
  \caption{Overview of the proposed APB2FaceV2 that consists of an Audio-aware Fuser $\boldsymbol{\phi}$ and a Multi-face Reenactor $\boldsymbol{\psi}$. $\boldsymbol{\phi}$ inputs audio, head pose, and eye blink signals that are extracted by $\boldsymbol{\phi}_A$, $\boldsymbol{\phi}_H$, and $\boldsymbol{\phi}_E$ to obtain $\boldsymbol{f}_A$, $\boldsymbol{f}_H$, and $\boldsymbol{f}_E$ respectively. Then these features are concatenated and extracted through $\boldsymbol{\phi}_D$ to obtain the embedded representation of the facial geometric feature $\boldsymbol{f}_{Geo}$. Subsequently, $\boldsymbol{\psi}$ inputs a reference face image $\boldsymbol{I}_{R}$ and the extracted facial geometric feature $\boldsymbol{f}_{Geo}$ to reenact the target face $\boldsymbol{\hat{I}}_{T}$ that has matched facial expression with the input signals. Specifically, a novel AdaConv is proposed to inject facial geometric information into the reenactor using a more flexible and efficient way.}
  \label{fig:architecture}
  \vspace{-10pt}
\end{figure*}
\section{Related Works}
\label{sec:rela}
\noindent\textbf{Generative Adversarial Networks}.
Since the concept of generative adversarial network was first proposed~\cite{GAN}, many excellent works have been proposed to generate photorealistic images. Generally, these methods mainly fall into two categories: the vector-based method~\cite{PGGAN, StyleGAN, StyleGAN2} that only uses a noise or embedded vector as input to generate target image, and the pixel-based method~\cite{Pix2Pix, CycleGAN} that uses the image as input. Theoretically, each of these methods contains a generator \emph{G} with parameter $\theta_{g}$ to capture the data distribution for generating photorealistic images, as well as a discriminator \emph{D} to authenticate generated images for enhancing the capability of \emph{G} in an adversarial manner. To learn the distribution of \emph{G} over data $x$ from a prior distribution $p{_z}(z)$ ($G(z; \theta_{g}) \in p_{data}(x)$), \emph{D} plays a two-player minimax game with \emph{G} in the following value function $V(D, G)$:

\begin{equation}
  \begin{aligned}
    \underset{G}\min~\underset{D}\max~V(D, G) &= \mathbb{E}_{\boldsymbol{x} \sim p_{data}(\boldsymbol{x})}[\log (D(\boldsymbol{x}))] \\
    &+ \mathbb{E}_{\boldsymbol{z} \sim p_{z}(\boldsymbol{z})}[\log (1 - D(G(\boldsymbol{z})))].
  \end{aligned}
  \end{equation}
Our proposed method belongs to the pixel-based category that inputs an image instead of only  a vector, which is more efficient and practical than vector-based method.

\noindent\textbf{Face Reenactment via Audio}. 
Many works have yielded good results in the audio-to-face task, which uses audio as input for providing adequate information about orofacial movements. Works~\cite{sadoughi2019speech, yi2020audio, cudeiro2019capture, Audio2face} use the audio signal to predict parameters of the predefined 3D model, while Suwajanakorn~\textit{et al.}~\cite{suwajanakorn2017synthesizing} and Prajwal~\textit{et al.}~\cite{prajwal2020lip} propose to predict the lip rather the full face. Thus these methods need extra post-operations such as 3D rendering or face fusion, which is cumbersome and not suitable for practical applications. We wish design an end-to-end method to directly generate the full face like~\cite{X2Face, Speech2face, chung2017you, bai2020speech, choi2020inference}, and supplement some auxiliary signals simultaneously to control the facial areas that are not related to the audio information. So based on the previous work~\cite{APB2Face}, we design a new framework named APB2FaceV2 that can not only generate photorealistic face end-to-end but also reenact multiple faces in real time by a unified model. 
\section{Method}
\label{sec:method}
As depicted in Figure~\ref{fig:architecture}, we propose a novel APB2FaceV2 framework, which consists of an \emph{Audio-aware Fuser} ($\boldsymbol{\phi}$) and a \emph{Multi-face Reenactor} ($\boldsymbol{\psi}$), to complete a more challengeable audio-guided multi-face reenactment task efficiently in real time. Detailed implementation and source code are available.

\noindent\textbf{Audio-aware Fuser}. The Audio-aware Fuser module inputs audio, head pose, and eye blink signals, which are further extracted by $\boldsymbol{\phi}_A$, $\boldsymbol{\phi}_H$, and $\boldsymbol{\phi}_E$ to obtain $\boldsymbol{f}_A$, $\boldsymbol{f}_H$, and $\boldsymbol{f}_E$ respectively. Specifically, $\boldsymbol{\phi}_A$ contains 5 convolutional layers for extracting the feature of each time node and additional 5 convolutional layers for fusing them, while both modules $\boldsymbol{\phi}_H$ and $\boldsymbol{\phi}_E$ contain three linear layers. Subsequently, the three features are concatenated and further extracted through $\boldsymbol{\phi}_D$ to obtain the embedded representation of the facial geometric feature $\boldsymbol{f}_{Geo}$, where the facial landmark is used as the supervisory signal in the training stage.
\begin{equation}
  \vspace{-3pt}
  \begin{aligned}
    \boldsymbol{f}_{Geo} = \boldsymbol{\phi}_{D}(\boldsymbol{f}_{Fus}) = \boldsymbol{\phi}_{D}([\boldsymbol{f}_A, \boldsymbol{f}_H, \boldsymbol{f}_E]),
  \end{aligned}
  \vspace{-3pt}
\end{equation}

\noindent\textbf{Multi-face Reenactor}. Given a reference face image $\boldsymbol{I}_{R}$ and the extracted facial geometric feature $\boldsymbol{f}_{Geo}$, the Multi-face Reenactor $\boldsymbol{\psi}$ reenacts the target face $\boldsymbol{\hat{I}}_{T}$ that has matched facial expression with the input signals, i.e. audio, head pose, and eye blink. Specifically, $\boldsymbol{\psi}$ consists of a chain of sub-modules: an image encoder $\boldsymbol{\psi}_E$, a feature transformer $\boldsymbol{\psi}_T = \left\{\boldsymbol{\psi}_T^1, \boldsymbol{\psi}_T^2, \dots, \boldsymbol{\psi}_T^L\right\}$ ($L$ represents the number of module repetitions and is set to 9 in the paper), and an image decoder $\boldsymbol{\psi}_D$. The process can be described as:
\begin{equation}
  \vspace{-3pt}
  \begin{aligned}
    \boldsymbol{\hat{I}}_{T} = \boldsymbol{\psi}_{D}(\boldsymbol{\psi}_{T}(\boldsymbol{\psi}_{E}(\boldsymbol{I}_{R}))),
  \end{aligned}
  \vspace{-3pt}
\end{equation}
Note that the feature transformer $\boldsymbol{\psi}_T$ is designed in a decoupling idea that simultaneously learns appearance information from $\boldsymbol{I}_{R}$ as well as the geometric information from $\boldsymbol{f}_{Geo}$, and the new proposed \emph{AdaConv} is used to inject geometric information on each block. 

\noindent\textbf{Adaptive Convolution}. Different from APB2Face~\cite{APB2Face} that injects facial movement information by first plotting the landmark image and then concatenating it with deep features, we propose an elegant information injection module, i.e. Adaptive Convolution (AdaConv). As shown in the top right of Figure~\ref{fig:architecture} (Highlighted in green), the AdaConv layer inputs a geometric vector $\boldsymbol{f}_{Geo}$ and a deep feature map $\boldsymbol{F}_{in}^{l}$, and outputs a modified feature map $\boldsymbol{F}_{out}^{l}$. In detail, $\boldsymbol{f}_{Geo}$ goes through two linear layers to generate a set of parameters, i.e. $\boldsymbol{f}_{Geo} \rightarrow \boldsymbol{W}^{l}$, which will be reshaped and applied to a convolutional layer. As the formula shows below:
\begin{equation}
  \vspace{-3pt}
  \begin{aligned}
    \boldsymbol{F}_{out}^{l} &= Conv(\boldsymbol{F}_{in}^{l}; \boldsymbol{W}^{l}) \\
    &= Conv(\boldsymbol{F}_{in}^{l}; Linear^l(\boldsymbol{f}_{Geo})),
  \end{aligned}
  \vspace{-3pt}
\end{equation}
The parameter $\boldsymbol{W}^{l}$ of the convolutional layer contains $k \times k \times C_g \times C$ weight parameters and $C$ bias parameters, where $k$, $C$, and $C_g$ are kernel size, channel number, and group number for the convolution. Thus we can control the amount of injected information by controlling the values of these parameters. Specifically, AdaConv reduces to AdaIN~\cite{AdaIN} when we set $\{k=1, C_g=1\}$, and we set $\{k=3, C_g=1\}$ in the paper for balancing computation and model effect. 

\noindent\textbf{Objective Function}. In the training stage, we adopt geometry and content losses to supervise geometric information and generated image quality, as well as adversarial loss to further improve the quality and authenticity of the reenacted image. The overall loss function is:
\begin{equation}
  \vspace{-3pt}
  \begin{aligned}
    \mathcal{L}_{All} = \lambda_{G} \mathcal{L}_{G} + \lambda_{C} \mathcal{L}_{C} + \lambda_{Adv} \mathcal{L}_{Adv},
  \end{aligned}
\end{equation}
where $\lambda_{G}$, $\lambda_{C}$, and $\lambda_{Adv}$ represent weight parameters to balance different terms, and are set 1, 100, and 1 respectively.

\romannumeral1) Geometry loss $\mathcal{L}_{G}$ calculates $\ell_1$ error between the predicted facial geometric feature $\boldsymbol{f}_{Geo}$ and corresponding real facial landmark $\boldsymbol{l}$.
\begin{equation}
  \begin{aligned}
    \mathcal{L}_{G} = || \boldsymbol{f}_{Geo} - \boldsymbol{l} ||_{1}.
  \end{aligned}
\end{equation}

\romannumeral2) Content loss $\mathcal{L}_{C}$ calculates $\ell_1$ error between the reenacted target face $\boldsymbol{\hat{I}}_{T}$ and corresponding real face $\boldsymbol{I}_{T}$.
\begin{equation}
  \begin{aligned}
    \mathcal{L}_{C} = || \boldsymbol{\hat{I}}_{T} - \boldsymbol{I}_{T} ||_{1}.
  \end{aligned}
\end{equation}

\romannumeral3) Adversarial loss $\mathcal{L}_{Adv}$ adopts an extra discriminator $D$ to form a adversarial training against the reenactor $G$ that greatly improves the quality of the generated image. 
\begin{equation}
  \begin{aligned}
    \mathcal{L}_{Adv} = \mathbb{E}_{\boldsymbol{\hat{I}}_{T} \sim p_{f}}[D({\boldsymbol{\hat{I}}_{T}})] - \mathbb{E}_{\boldsymbol{I}_{T} \sim p_{r}}[D(\boldsymbol{I}_{T})],
  \end{aligned}
\end{equation}
where $p_{r}$ and $p_{f}$ stand distributions for real and generated fake images respectively.
\vspace{-5pt}
\section{Experiments}
\label{sec:exp}

\noindent\textbf{Dataset}. In the paper, almost all of experiments are conducted on AnnVI dataset that contains six announcers (three men and three women) and 23790 frames totally with corresponding audio clip, head pose, eye blink, and landmark~\cite{Face}.

\noindent\textbf{Implementation Details}. We use Adam optimizer~\cite{Adam} with $\{\beta_1=0.5$, $\beta_2=0.999\}$ and train the model for 110 epoch. The learning rate is set to $2e^{-4}$, and the batch size is 16. PatchGAN~\cite{Pix2Pix} is used as the discriminator, and the training setting is in accord with the reenactor. 

\begin{figure}[!ht]
  \centering
  \includegraphics[width=1\columnwidth]{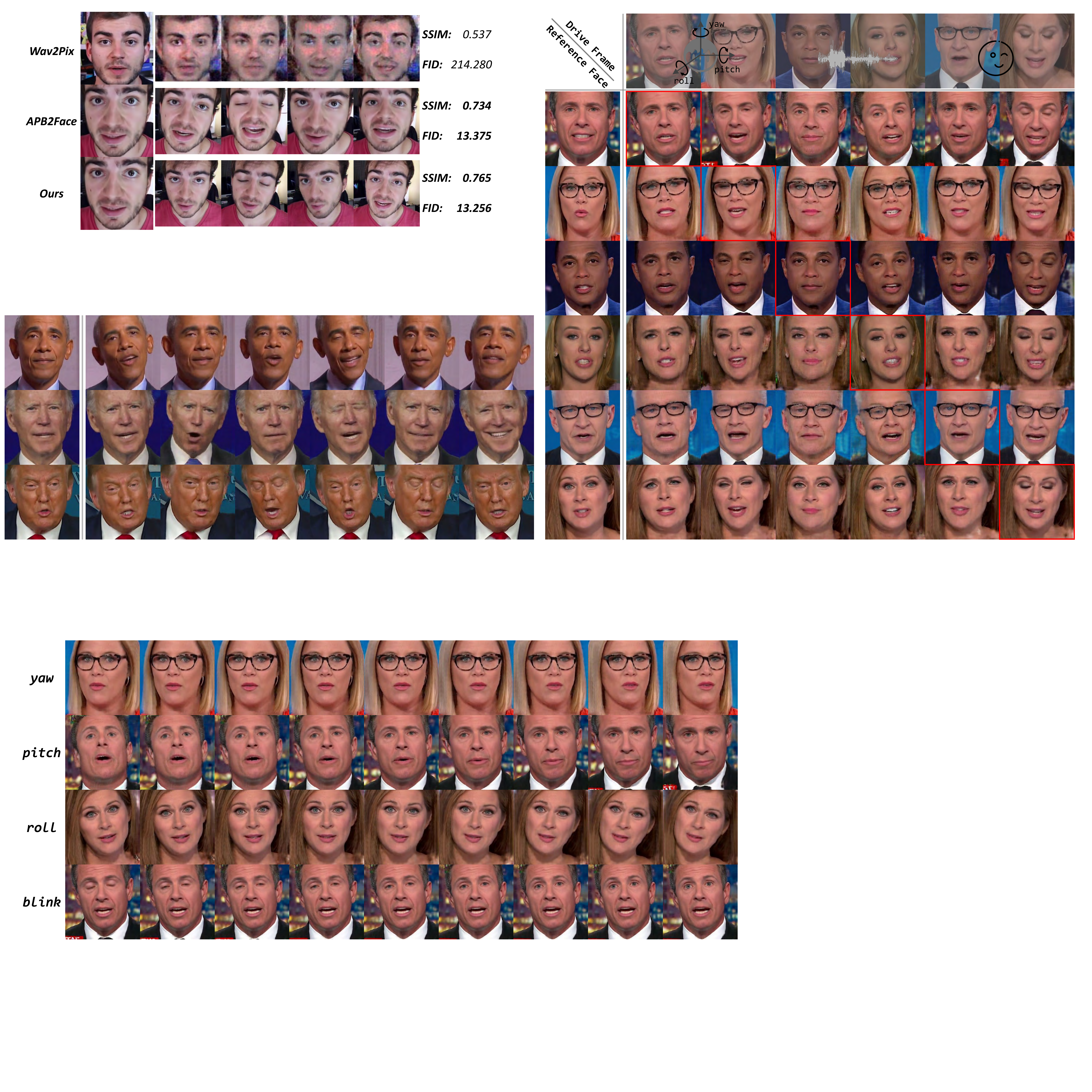}
  \caption{Experimental results among multiple persons on the AnnVI dataset. The first column contains six randomly selected reference faces, and the first row shows driven frames from different persons that supply audio, head pose, and eye blink signals. The generated faces marked with red rectangles are driven by the personal own audio, while other faces are driven by various persons for evaluating the generalization ability of the network.}
  \label{fig:generation}
  \vspace{-5pt}
\end{figure}


\noindent\textbf{Qualitative Results}. Some qualitative experiments are conducted on AnnVI dataset to visually demonstrate the high quality of reenacted images and the flexibility of the proposed approach. Specifically, we randomly select one face of each identity (6 faces totally) as the reference face, and one drive frame of each identity for supplying audio, head pose, and eye blink signals. As indicated by the red rectangles in Figure~\ref{fig:generation}, our proposed method can reenact photorealistic faces among multiple persons with one unified model that achieves multi-face reenactment task. Thanks to the decoupling design of our method, APB2FaceV2 can use input signals of other persons to reenact the target face that is consistent with the identity of the reference face. Experimental results show that our method has strong generalization ability, where the model can use non-self audio as input to reenact photorealistic faces.

\noindent\textbf{Quantitative Results}. As shown in Table~\ref{tab:metrics}, SSIM metric is chosen to quantitatively evaluate our method with the state-of-the-art (SOTA) method, and experimental results indicate that the proposed method can generate more photorealistic faces where the SSIM score goes up from 0.799 to 0.805, even though using only one unified model (The work~\cite{APB2Face} need train 6 models totally for 6 persons). Nevertheless, our method still obtains a higher \emph{Detection Rate} (DR), i.e. 99.1\%, which also demonstrates the superiority of our method than SOTA.

\begin{table}[t] \small
  \begin{center}
  \caption{Quantitative assessment for comparing our method with the SOTA APB2Face~\cite{APB2Face} in SSIM~\cite{SSIM} and DR~\cite{APB2Face} of different persons on the AnnVI dataset.}
  \label{tab:metrics}
  \resizebox{240pt}{26pt}{
    \begin{tabular}{cccccccc}
    \noalign{\smallskip}
    Metric & $P1$ & $P2$ & $P3$ & $P4$ & $P5$ & $P6$ & $Avg$\\
    \hline
    SSIM~(\cite{APB2Face}) $\uparrow$ & 0.764 & 0.843 & 0.879 & 0.786 & 0.761 & 0.758 & 0.799 \\
    SSIM~(Ours) $\uparrow$& \textbf{0.758} & \textbf{0.860} & \textbf{0.862} & \textbf{0.809} & \textbf{0.790} & \textbf{0.752} & \textbf{0.805} \\
    \hline
    DR~(\cite{APB2Face}) $\uparrow$& 98.9 & 98.8 & 99.1 & 98.8 & 98.7 & 98.5 & 98.8 \\
    DR~(Ours) $\uparrow$& \textbf{99.1} & \textbf{99.0} & \textbf{99.4} & \textbf{99.3} & \textbf{98.7} & \textbf{98.8} & \textbf{99.1} \\
    \hline
    \end{tabular}
    }
  \end{center}
  \vspace{-10pt}
\end{table}

\begin{figure}[!ht]
  \centering
  \includegraphics[width=1\columnwidth]{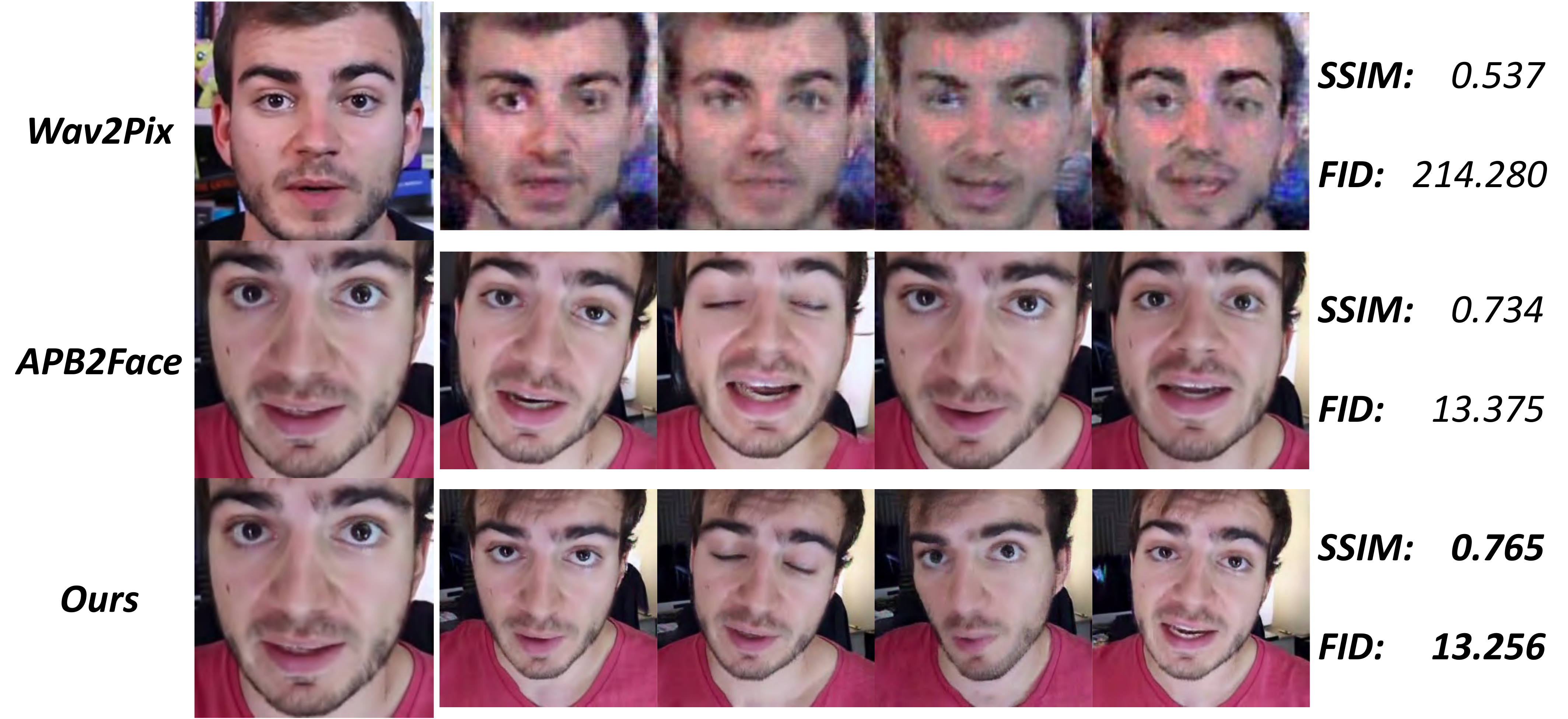}
  \caption{Qualitative comparison experiment with SOTA Wav2Pix~\cite{Wav2Pix} and APB2Face~\cite{APB2Face}.}
  \label{fig:comparison}
  \vspace{-10pt}
\end{figure}

\begin{table}[t] \small
  \begin{center}
  \caption{Qualitative comparison experiment with SOTAs in several metrics on the AnnVI dataset.}
  \label{tab:comparison}
  \resizebox{240pt}{21pt}{
    \begin{tabular}{cccccccc}
    \noalign{\smallskip}
    Method & SSIM $\uparrow$ & FID $\downarrow$ & Params (M) & FPS (CPU) & FPS (GPU) \\
    \hline
    Wav2Pix~\cite{Wav2Pix}   & 0.537 & 214.280 & 24.771 & 20.7 & \textbf{275.7} \\
    APB2Face~\cite{APB2Face} & 0.734 & 13.375  & 16.696 & 5.8  & 200.4 \\
    Ours                     & \textbf{0.765} & \textbf{13.256}  & \textbf{4.085}  & \textbf{22.5} & 158.9 \\
    \hline
    \end{tabular}
    }
  \end{center}
  \vspace{-10pt}
\end{table}

\noindent\textbf{Comparison with SOTAs}. We further conduct a comparison experiment with most related SOTA methods on the Youtubers dataset~\cite{Wav2Pix}. As show in Figure~\ref{fig:comparison}, our approach obtains a better visual effect than others, as well as the best SSIM and FID scores. Moreover, our method significantly reduces the number of parameters by nearly 6 times and 4 times compared to Wav2Pix and APB2Face respectively, and can run in real time, i.e. \textbf{22.5} FPS in CPU (i7-8700K @ 3.70GHz) and \textbf{158.9} FPS in GPU (2080 Ti), as shown in Table~\ref{tab:comparison}.

\begin{figure}[!ht]
  \vspace{10pt}
  \centering
  \includegraphics[width=1\columnwidth]{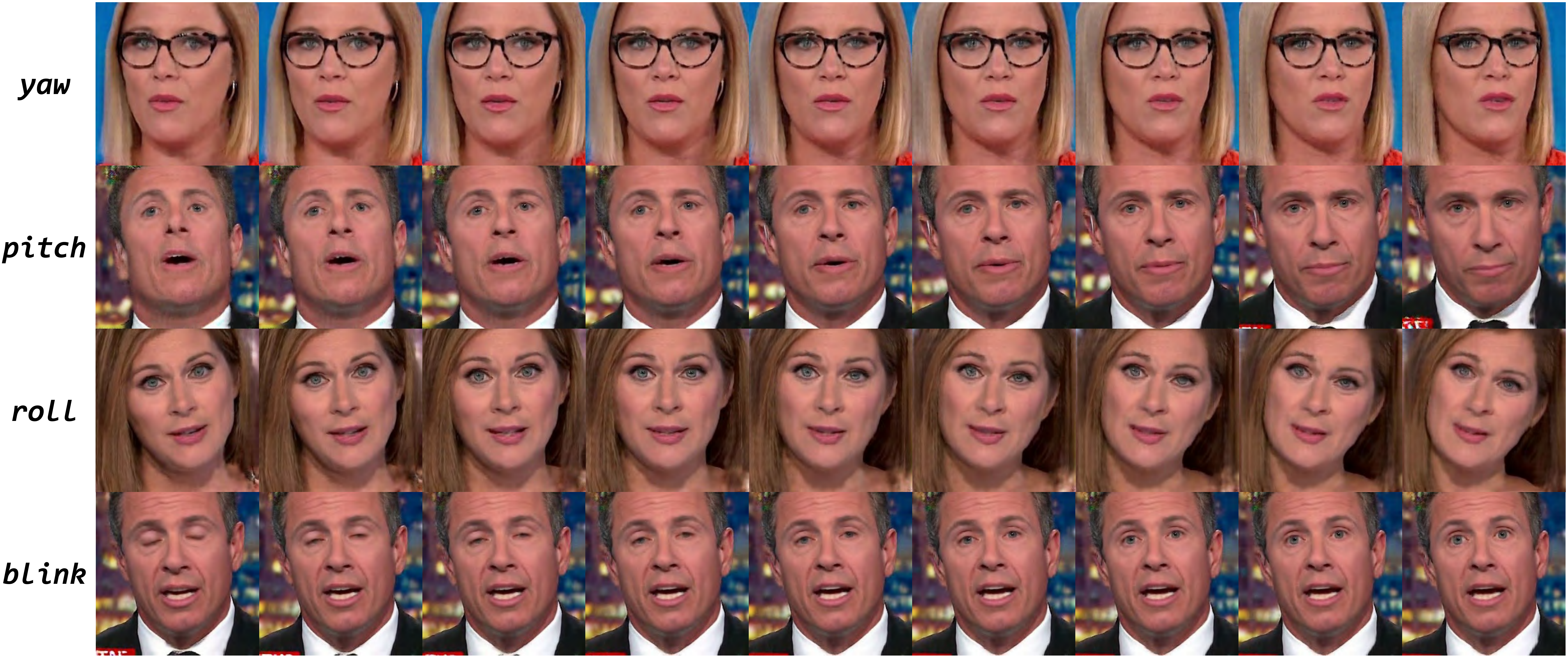}
  \caption{Decoupling experiments of pose and eye blink signals.}
  \label{fig:decoupling}
  \vspace{-5pt}
\end{figure}

\noindent\textbf{Decoupling Experiment}. A decoupling experiment is further conducted to demonstrate that our proposed method is capable of disentangling input signals, i.e. audio, head pose, and eye blink. As shown in Figure~\ref{fig:decoupling}, the first three rows are generated results that only change one component of the head pose signal, i.e. yaw, pitch, or roll, while the last row shows the results that only change the eye blink signal. Experimental results indicate that our method can control the properties of the generated face that is flexible for practical applications.
\section{Conclusions}
\label{sec:con}
In this paper, we propose a novel APB2FaceV2 to address a more challengeable audio-guide multi-face reenactment task, which aims at using one unified model to reenact different target faces among multiple persons with corresponding reference face and drive audio signal as inputs. Specifically, an Audio-aware Fuser is firstly used to predict a geometric representation from input signals, and then Multi-face Reenactor fuse it with the reference face that supplies appearance information to reenact photorealistic target face. Besides, a novel AdaConv module is proposed to inject geometric information in a more elegant and efficient way. Extensive experiments demonstrate the efficiency and flexibility of our approach.

We will further combine \emph{Neural Architecture Search} (NAS) with our approach to search for a more accurate and faster model for better practical applications, and we hope our work will help users to achieve better jobs.

\bibliographystyle{IEEEbib}
\bibliography{refs}

\end{document}